\documentclass[superscriptaddress,twocolumn,prl,amsmath,amssymb,floatfix]{revtex4-2}

\usepackage[pdftex]{graphicx}
\usepackage{dcolumn}
\usepackage{bm}

\begin{document}
\title{Designing Chaotic Attractors:\\
    A Semi-supervised Approach
}

\author{Tempei Kabayama}
\author{Yasuo Kuniyoshi}%
\affiliation{
 Graduate School of Information Science and Technology, University of Tokyo, Bunkyo-ku, Tokyo 113-0033, Japan
}
\author{Kazuyuki Aihara}
\affiliation{
 International Research Center for Neurointelligence, University of Tokyo Institutes for Advanced Study, University of Tokyo, Bunkyo-ku, Tokyo 113-0033, Japan
}
\author{Kohei Nakajima}
\affiliation{
 Graduate School of Information Science and Technology, University of Tokyo, Bunkyo-ku, Tokyo 113-0033, Japan
}

\date{\today}

\begin{abstract}
Chaotic dynamics are ubiquitous in nature and useful in engineering, but their geometric design can be challenging. Here, we propose a method using reservoir computing to generate chaos with a desired shape by providing a periodic orbit as a template, called a \textit{skeleton}. We exploit a bifurcation of the reservoir to intentionally induce unsuccessful training of the \textit{skeleton}, revealing inherent chaos. The emergence of this untrained attractor, resulting from the interaction between the \textit{skeleton} and the reservoir's intrinsic dynamics, offers a novel semi-supervised framework for designing chaos.
\end{abstract}

\pacs{Valid PACS appear here}

\maketitle

Chaotic dynamics are prevalent in nature, including biological neural systems \cite{hayashi1982chaotic, aihara1986structures, freeman1987simulation}, and are applied in engineering, 
such as for random number generation \cite{akashi2022mechanical, uchida2008fast}, communication systems \cite{argyris2005chaos, boccaletti2000control}, optimization \cite{chen1995chaotic, kong2020stochasticity}, deep learning \cite{inoue2022transient, liu2024exploiting}, and robot control \cite{namikawa2011neurodynamic, laje2013robust, inoue2020designing}.

\begin{figure}[!t]
    \begin{center}
     \includegraphics[scale=1.0]{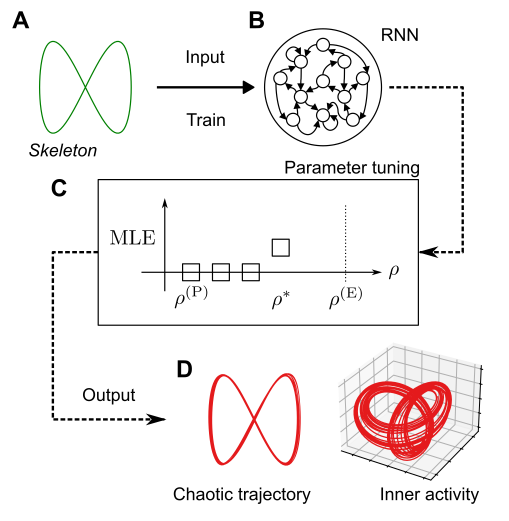}
     \caption{
     Schematic diagram of the proposed method.
     (A) Periodic time series defining the overall shape.
     (B) Reservoir learning the \textit{\textit{skeleton}}.
     (C) Parameter exploration: Finding the parameter $\rho^{*}$.
     (D) Designed chaotic orbit. Left: Reservoir's output. Right: Plot of the internal state based on the principal components.
     }
     \label{fig:scheme}
    \end{center}
\end{figure}

A notable challenge is designing the geometric shapes of chaotic attractors, for which no practical methods have been proposed so far, to the best of our knowledge. 
In this paper, as shown in Fig.~\ref{fig:scheme}, we propose a method using the reservoir computing (RC) framework \cite{jaeger2001echo,maass2002real,nakajima2021reservoir}.
In this method, a periodic orbit, which we call a \textit{skeleton}, defines the contour of the output trajectory. 
Training the reservoir with the \textit{skeleton} under certain parameter values yields an autonomous system, where the maximum Lyapunov exponent (MLE) exceeds zero, while the trajectory mimics the shape of the \textit{skeleton}, indicating periodic chaos.

The proposed method utilizes attractor reconstruction with RC \cite{lu2018attractor,hart2020embedding}, which is achieved through one-step-ahead prediction of the target inputs and closed-loop operation.
The premise of RC is to use a fixed nonlinear dynamical system of the following form:
\begin{subequations}
    \begin{align}
        \bm{x}_{k+1} &= F(\bm{x}_{k}, \bm{u}_{k})\label{eq:res_ds},\\
        \bm{z}_{k} &= \bm{W}^{\top}_{out} \bm{x}_{k}\label{eq:res_ro}.
    \end{align}
\end{subequations}
Here, the reservoir \eqref{eq:res_ds} is an $N$-dimensional system with a $D$-dimensional input, where $\bm{x}_{k}\in\mathbb{R}^{N}$ and $\bm{u}_{k}\in\mathbb{R}^{D}$ represent the reservoir state and the input, respectively. 
The output $\bm{z}_{k}$ is defined as a linear combination of $\bm{x}_{k}$.
Typically, recurrent neural networks (RNNs) are employed as reservoirs.
The readout layer $\bm{W}_{out}\in\mathbb{R}^{N \times D}$ is constructed so that $\bm{z}_{k}\simeq\bm{u}_{k}$ using ridge regression with following process:
the reservoir is driven by the teacher input (teacher forcing), and the set of the response is obtained for $T_{\mathrm{init}}+T_{\mathrm{train}}$ steps, with the initial $T_{\mathrm{init}}$ steps discarded. 
This yields the paired learning data:
\begin{equation}
        \bm{X} = \left[\bm{x}_{T_{\mathrm{init}}} \cdots \bm{x}_{T_{\mathrm{init}}+T_{\mathrm{train}}}\right],
        \bm{Y} = \left[\bm{u}_{T_{\mathrm{init}}} \cdots \bm{u}_{T_{\mathrm{init}}+T_{\mathrm{train}}}\right].
\end{equation}
Then, $\bm{W}_{out}$ is determined by
\begin{equation}
    \bm{W}_{out} = (\bm{X}^{\top}\bm{X} + \beta \bm{I})^{-1}\bm{X}^{\top}\bm{Y},
\end{equation}
where $\beta$ is the regularization parameter.
By feeding back the predicted output $\bm{z}_{k}$ as the alternative input, the system acquires autonomous dynamics:
\begin{equation}
    \bm{x}_{k+1} = F(\bm{x}_{k}, \bm{z}_{k}) = \hat{F}(\bm{x}_{k}),
\end{equation}\label{eq:res_cl}
where the initial state is the final state of the teacher forcing.
The well-trained system replicates the target attractor, mimicking its ergodic characteristics and providing estimates of dynamical quantities, such as Lyapunov exponents (LE) \cite{lu2018attractor, pathak2017using}. 
Conversely, this autonomous system may exhibit some properties not present in the target, termed untrained attractors \cite{lu2018attractor, flynn2021multifunctionality}.

The attractor shown in Fig.~\ref{fig:scheme}D is a chaotic untrained attractor.
We demonstrate that the emergence of such attractors can be induced by appropriate parameter selection in the learning periodic sequences and formalize this fact as a method for designing chaotic attractors.
Importantly, the obtained chaos is formed by integrating the teacher information and the intrinsic dynamical properties of the reservoir.
Therefore, this method can be considered a novel ``semi-supervised" framework.

As a case study on learning periodic orbits, we consider the Lissajous curve $\bm{u}_k=\left[\cos \tfrac{\pi k}{50}, \sin \frac{\pi k}{25}\right]^\top$ shown in Fig.~\ref{fig:scheme}A.
In this paper, the reservoir is a leaky integrator echo state network (LESN) \cite{jaeger2007optimization}, described by
\begin{equation}
    \bm{x}_{k+1} = (1-a)\bm{x}_{k} + a \tanh(\rho \bm{W} \bm{x}_{k} + \sigma \bm{W}_{in} \bm{u}_{k}), \label{eq:lesn}
\end{equation}
where $a$ is the leaky rate, and $\tanh$ is an element-wise function.
Here, $\bm{W}\in\mathbb{R}^{N \times N}$ is a random matrix whose elements follow a standard normal distribution, and it is scaled to have a spectral radius at unity. 
Thus, $\rho$ is the spectral radius of the internal connection matrix $\rho \bm{W}$. 
The input layer $\bm{W}_{in}\in\mathbb{R}^{N \times D}$ is a random matrix whose elements follow a uniform distribution in  $[-1, 1]$, and $\sigma$ represents the input intensity.

The dynamical properties of a random RNN without input depend largely on the spectral radius of its internal connection matrix \cite{sompolinsky1988chaos, cessac2007neuron}.
For an LESN, we introduce the effective spectral radius \cite{jaeger2007optimization} as follows:
\begin{equation}
    \rho_{e}^{(\mathrm{pre})} = |\lambda|_{\mathrm{max}}(a \rho \bm{W} + (1-a) \bm{I}).
\end{equation}
Typically, the MLE of the input-free LESN increases with $\rho_{e}^{(\mathrm{pre})}$ and exceeds zero at unity, as shown in Fig.~S1 in the Supplementary Material. 
To ensure reservoir convergence, Jaeger \textit{et al.} \cite{jaeger2007optimization} recommend keeping $\rho_e<1$ in practice. 
However, chaos in RNNs can be suppressed by external inputs \cite{molgedey1992suppressing}. 
Even with $\rho_e>1$, successful learning is feasible if the input-driven reservoir exhibits convergence properties, which is guaranteed by the negative conditional Lyapunov exponent (CLE) \cite{verstraeten2007experimental, lu2018attractor}.

\begin{figure}[tb]
    \begin{center}
     \includegraphics[scale=1.0]{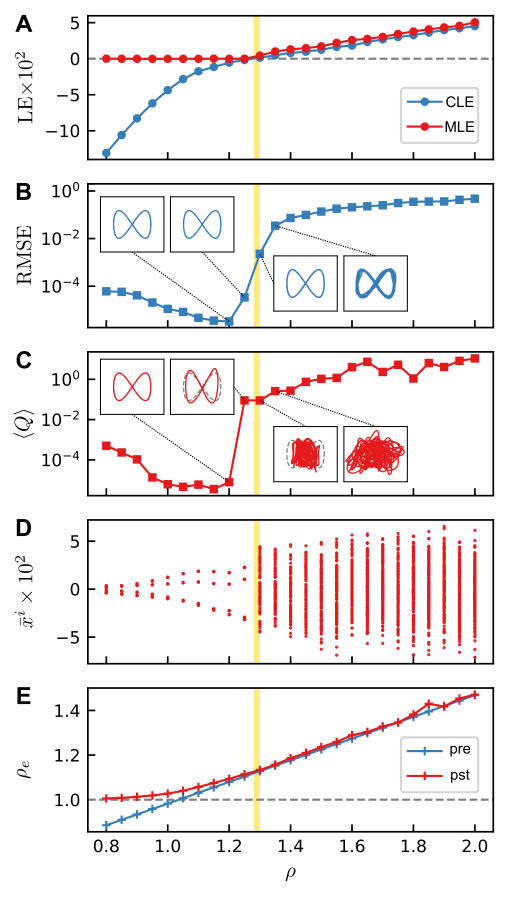}
     \caption{
     Analysis of learning the Lissajous curve.
     (A) CLE of the driven system and MLE of the closed-loop system.
     (B) RMSE in $x$ component in open-loop prediction for 10{,}000 steps and predictions at $\rho\in\{1.20, 1.25, 1.30, 1.35\}$.
     (C) Time average of evaluation index $Q$ for the output of the closed-loop system over 2{,}000 steps, and trajectories at the corresponding points.
     (D) Bifurcation diagram based on extrema of node averages $\bar{x}^i_k=\frac{1}{N}\sum_{i=1}^{N}x^i_k$.
     Calculated for 10{,}000 steps, discarding the first 8{,}000 steps.
     (E) Pre- and post-training effective spectral radius labelled ``pre" and ``pst," respectively.
     }
     \label{fig:macro}
    \end{center}
\end{figure}

As shown in Fig.~\ref{fig:macro}A, when the parameters are fixed at $(N, a, \sigma)=(1000, 0.5, 0.2)$, the CLE with Lissajous curve input increases with $\rho$ and the sign changes in $\rho\in[1.25, 1.30]$. This region corresponds to the ``edge of chaos" of the driven system, indicating the limit of chaos suppression \cite{haruna2019optimal}. 
In this paper, the range of $\rho$ with positive CLEs are referred to as the ``chaotic region," and those with negative CLEs as the ``convergent region," respectively.

As shown in Fig.~\ref{fig:macro}B,
the performance of the open-loop prediction was evaluated using the root mean square error, $\mathrm{RMSE}=\sqrt{\frac{1}{T_{\mathrm{eval}}}\sum_{k=1}^{T_{eval}}(z^{(d)}_k-u^{(d)}_{k})^2}$.
The prediction can be considered successful when $\mathrm{RMSE}<10^{-2}$. The range $\rho\in[0.8, 1.3]$ meets this criterion, with CLEs being negative or close to zero. 
The error decreases until $\rho=1.2$ and then increases.
Although the prediction is successful for $\rho=1.25$ and $\rho=1.30$, the excessively slow convergence of the driven reservoir likely negatively impacts the task. 
Additionally, a negative CLE does not guarantee a successful prediction and reconstruction. 
They also depend on the reservoir's short-term memory and information processing capacity, which can be influenced by $\rho$ \cite{haruna2019optimal}. 
Thus, the range of successful learning for $\rho$ depends on the selected target.

The trained closed-loop model is described by
\begin{equation}
    \begin{aligned}
        \bm{x}_{k+1} &= (1-a)\bm{x}_{k} + a \tanh(\left[\rho \bm{W} + \sigma \bm{W}_{in} \bm{W}^{\top}_{out}\right]\bm{x}_{k})\\
                     &= (1-a)\bm{x}_{k} + a \tanh(\hat{\bm{W}}\bm{x}_{k}).
    \end{aligned}
    \label{eq:lesn_cl}
\end{equation}
Here, the learning can be understood as a change in the matrix from $\rho\bm{W}$ to $\hat{\bm{W}}$ by adding the readout matrix \cite{sussillo2012transferring}. We fix $\beta=0.001$ and compare the properties of the closed-loop model for each $\rho$. 
To evaluate the shape of the output, we introduce an indicator $Q(x, y) = \left|x^4 - x^2 + \frac{1}{4}y^2\right|$, which is obtained by eliminating the time variable from $\bm{u}_k$.
This index is zero on the Lissajous curve, with a larger value indicating greater deviation. 
Here, we use the time average $\langle Q(z^{(x)}_k, z^{(y)}_k) \rangle$. 
If $\langle Q \rangle<10^{-2}$, the trajectory is ``along" the target curve.
As shown in Figs.~\ref{fig:macro}A and \ref{fig:macro}C, in the chaotic region ($1.3 \leq \rho$), the MLEs have positive values, and the learning fails. 
In contrast, in the convergence region ($\rho \leq 1.25$), the MLEs are zero, indicating a periodic attractor. 
Especially for $\rho \leq 1.2$, trajectories with $\langle Q \rangle<10^{-2}$ are obtained. 

Figure~\ref{fig:macro}D displays the extrema of the node averages $\bar{x}^i_k$ of the internal states $\bm{x}_k=[x^0_k, \ldots, x^N_k]^{\top}$ in the last 2{,}000 steps of the 10{,}000-step trajectory for each $\rho$.
Because the post-training matrix $\hat{\bm{W}}$ is a function of $\rho$, Fig.~\ref{fig:macro}D can be interpreted as a bifurcation diagram of the autonomous system \eqref{eq:lesn_cl} with respect to the bifurcation parameter $\rho$, showing the parametric continuity of the periodic orbits and the transition to chaos.

Additionally, we introduce the post-training effective spectral radius:
\begin{equation}
    \rho_{e}^{(\mathrm{pst})} = |\lambda|_{\mathrm{max}}(a \hat{\bm{W}} + (1-a) \bm{I}).
\end{equation}
As shown in Fig.~\ref{fig:macro}E, $\rho_{e}^{(\mathrm{pst})}$ is clearly larger than $\rho_{e}^{(\mathrm{pre})}$ for $\rho<1.25$. 
Approaching the edge of chaos, $\rho_{e}^{(\mathrm{pst})}$ converges to $\rho_{e}^{(\mathrm{pre})}$, and for $\rho>1.3$, $\rho_{e}^{(\mathrm{pst})}$ is nearly equal to $\rho_{e}^{(\mathrm{pre})}$. 
Here, it can be inferred that in the convergence region where $\rho_e$ clearly increases, learning modifies the convergence property of the reservoir, realizing the periodic attractor as a ``supervised" attractor. 
Conversely, in the chaotic region, this operation fails, and the reservoir's chaos is foregrounded, inducing an untrained chaotic attractor, even if the open-loop prediction roughly succeeds. 

\begin{figure}[tb]
    \begin{center}
     \includegraphics[scale=1.0]{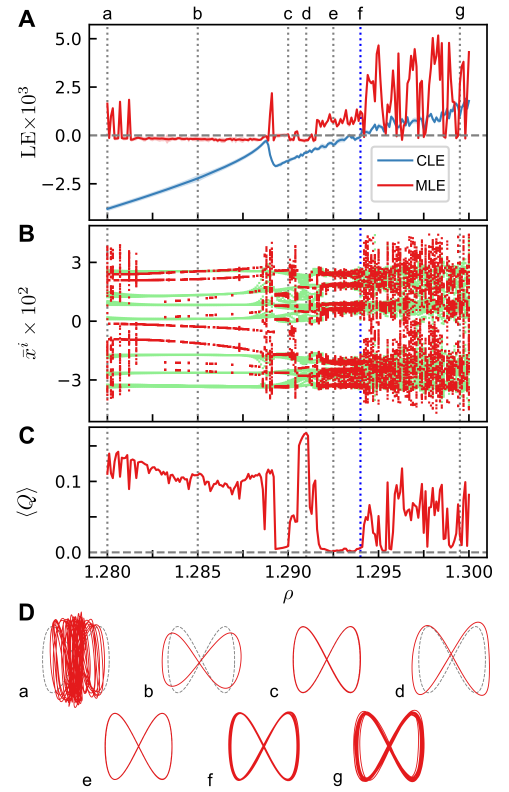}
     \caption{Analysis of learning the Lissajous curve near the edge of chaos (the region meshed in Fig.~\ref{fig:macro}). The blue dotted line f indicates $\rho^{(\mathrm{E})}$. (A) CLE and MLE. (B) Bifurcation diagram. Light green represents transients up to 2{,}000 steps. (C) Evaluation index $\langle Q \rangle$ in the closed-loop state. (D) Outputs of the autonomous system at points $\rho \in \{1.2800, 1.2850, 1.2900, 1.2910, 1.2925, 1.2940, 1.2995\}$ indicated by the dotted lines a--g.
     }
     \label{fig:edge}
    \end{center}
\end{figure}
We closely explore the vicinity of the edge of chaos, $\rho\in[1.28, 1.30]$, expecting a bifurcation structure connecting the supervised periodic orbits with the deformed chaotic attractors.
As shown in Fig.~\ref{fig:edge}A, the sign of CLE reverses at $\rho=1.2940$.
We denote this point as $\rho^{(\mathrm{E})}$ and focus on $\rho<\rho^{(\mathrm{E})}$, where the learning results are unstable.
As shown in Fig.~\ref{fig:edge}A, windows of periodic solutions with $\mathrm{MLE}=0$ alternate with chaotic regions with $\mathrm{MLE}>0$.
Figure~\ref{fig:edge}B is a bifurcation diagram obtained similarly to Fig.~\ref{fig:macro}D, but it also includes the first 2{,}000 steps. 
The trajectories appearing as continuous light green curves in the range $\rho \in [1.280, 1.289]$ correspond to the long transient on $\langle Q \rangle \simeq0$, implying that the Lissajous curve exists as an unstable orbit (Figs.~\ref{fig:edge}Da and \ref{fig:edge}Db). This orbit stabilizes (Fig.~\ref{fig:edge}Dc) and destabilizes again (Fig.~\ref{fig:edge}Dd).
In the region up to $\rho^{(\mathrm{E})}$ after the untrained periodic orbit destabilizes, we observe the chaotic attractors with $\mathrm{MLE}>0$ and $\langle Q \rangle \simeq0$ (Figs.~\ref{fig:edge}De and \ref{fig:edge}Df). 

These chaotic attractors appear as multiple separated bands in Fig.~\ref{fig:edge}B, indicating periodic/band chaos in terms of typical discrete-time dynamical systems.
They likely appear due to period-doubling bifurcations. 

The instability of results in this region can be attributed to the slow convergence of the driven system.
Therefore, the length of the truncated transient, $T_{\mathrm{init}}$, can act as a bifurcation parameter, which is discussed in the Supplementary Material.

Above $\rho^{(\mathrm{E})}$, the bands collapse, suggesting that a crisis occurs at this point. It is worth noting that even at $\rho>\rho^{(\mathrm{E})}$, a similar periodic chaos is found, for example, at $\rho=1.2995$, as shown in Fig.~\ref{fig:edge}Dg.
However, because the chaos in the driven system is not suppressed, the bifurcation structure in this region should be more complex than that for $\rho<\rho^{(\mathrm{E})}$.

From the above result, chaotic attractors with shapes along the Lissajous curve can be seen in the bifurcation process, in which the reconstruction of the periodic orbits collapses with the emergence of untrained chaotic attractors.
We refer to the points where supervised periodic orbits are reconstructed as the ``supervised points" $\rho^{(\mathrm{P})}$ and the points where chaotic attractors with shapes along the \textit{skeleton} emerge as the ``semi-supervised points" $\rho^{*}$.

In the trials using the Lissajous curve, $\rho^{*}$ can be found between $\rho^{(\mathrm{P})}$ and $\rho^{(\mathrm{E})}$ with different realizations of the random matrices $\bm{W}$ and $\bm{W}_{in}$.
It also holds when the \textit{skeleton} is changed, as shown in Fig.~S4 in the Supplementary Material.

Figure~\ref{fig:example}A shows the chaotic attractor generated with the Van der Pol oscillator.
It is worth noting that the time series is classified as a sequence of quasi-periodic orbits in discrete time, while the Lissajous curve is periodic orbits. Despite this difference, consistent results were obtained.
Furthermore, Fig.~\ref{fig:example}B shows the results using the hand-drawn closed curve, demonstrating that similar phenomena can be observed even when no explicit equations are known for the \textit{skeleton}.

The generation of semi-supervised points can be explained from the following two perspectives.
First, as previously discussed, fully supervised attractors are obtained with sufficiently fast convergence, while in chaotic regions, the reservoir chaos is foregrounded. Given a bifurcation structure by $\rho$ linking these two points, it is plausible that attractors in the intermediate region may display properties midway between the two.
Specifically, the generated chaos is thought to arise from the supervised periodic orbits becoming chaotic through a cascade of period-doubling bifurcation. 
Therefore, one trivial requirement is the existence of $\rho^{(\mathrm{P})}$.

Second, the CLE is crucial. 
According to Hart \cite{hart2024attractor}, the LEs of the well-trained reservoir reconstructing a chaotic system provides reliable estimates of the target's LEs only for those larger than the driven reservoir's CLE.
Negative LEs below the CLE tend to approximate those of the driven system rather than the ground truth.
Therefore, accurate reconstruction requires a sufficiently negative CLE.
Although we focus on periodic orbits, this fact is important. For example, when embedding a limit cycle, the MLE should be zero, and other exponents should be negative. If the CLE is zero, even with successful prediction, the non-convergence of the driven system makes the orbit unstable in the closed-loop state.

Moreover, the prediction always involves some errors, which are fed back into the reservoir and act as noise in subsequent predictions.
The accumulation of this noise determines the qualitative difference between the open-loop attractor $\Lambda_c$ and the closed-loop attractor $\Lambda$.
If the prediction is successful, the evolution of $\Lambda$ approximates $\Lambda_c$ with small noise.
Fast convergence of the driven system prevents error accumulation, resulting in $\Lambda \sim \Lambda_c$. 
Conversely, large prediction errors or slow convergence may lead to error accumulation and destabilize the orbit of $\Lambda_c$.
Therefore, it is plausible that the behavior of trajectories near the \textit{\textit{skeleton}} in a closed-loop state is determined by the balance between prediction errors and the convergence of the driven system. 
Thus, by appropriately selecting $\rho$ to determine the CLE, generating chaos near the \textit{\textit{skeleton}} can be achieved. 
However, the exploration of the corresponding parameters is heuristic.

As shown in Fig.~\ref{fig:scheme}, the procedure for identifying the semi-supervised points $\rho^*$ is formulated as follows:
\begin{enumerate}
    \item Prepare a periodic time series $\bm{u}_k$ as the \textit{skeleton} that defines the shape of the output.
    \item Prepare the reservoir with fixed parameters $(N, a, \sigma, \beta)$.
    \item Roughly identify $\rho^{(\mathrm{E})}$ in this setting.
    \item Repeat learning for $\rho<\rho^{(\mathrm{E})}$ to discover $\rho^{(\mathrm{P})}$.
    If not found, return to step 2 and modify the fixed setting.
    \item Explore the interval $[\rho^{(\mathrm{P})}, \rho^{(\mathrm{E})}]$ 
\end{enumerate}
Step 4 can be simplified using a binary search.
Step 5 essentially repeats learning and manually evaluates the results, but automation is possible if a geometric evaluation index like $Q$ is available. 
However, as shown in Fig.~\ref{fig:edge}, there may be multiple windows of periodic solutions, so a somewhat detailed search is required.

\begin{figure}[tb]
    \begin{center}
     \includegraphics[scale=1.0]{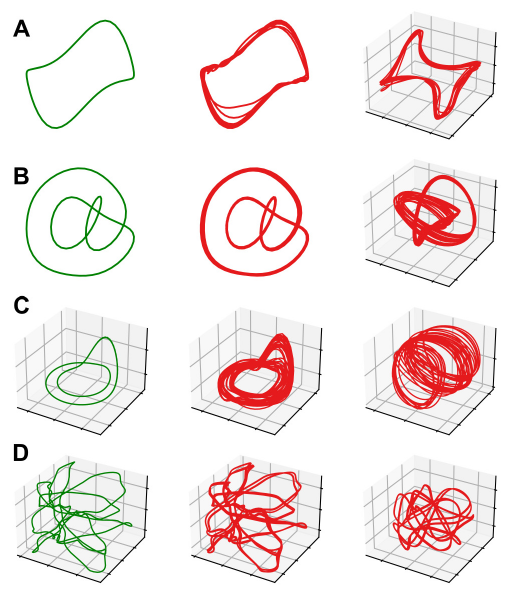}
     \caption{
     Examples of chaotic untrained attractors.
     The left panel shows \textit{skeleton} $\bm{u}_k$, the middle panel shows $\bm{z}_k$, and the right panel shows the plot based on the first and second principal components of $\bm{x}_k$.
     (A) Van der Pol equation. $\mathrm{MLE}=0.002$
     (B) Handwritten ``at" symbol. $\mathrm{MLE}=0.004$.
     (C) Periodic orbit of the R\"{o}ssler system. $\mathrm{MLE}=0.003$.
     (D) Time series created from the C-E-G keys of the piano. $\mathrm{MLE}=0.0003$.
     In all cases, $(a, \beta)=(0.5, 0.001)$.
     For A and B, $(N, \sigma)=(1000, 0.2)$;
     for C, $(1000, 0.02)$;
     and for D, $(1500, 0.2)$.
     The realizations of $\bm{W}$ are the same for A, B and C. For A and B, the realizations of $\bm{W}_{in}$ are also the same.
     }
     \label{fig:example}
    \end{center}
\end{figure}

The application results for three-dimensional data are shown in Fig.~\ref{fig:example}C and \ref{fig:example}D.
The trial shown in Fig.~\ref{fig:example}C uses a limit cycle obtained from the R\"{o}ssler system, and some of the obtained attractors resemble the chaotic attractors seen in the original system, which is discussed in the Supplementary Material.
The \textit{skeleton} in Fig.~\ref{fig:example}D is constructed by repeating sound waves produced by three keys on a piano.
The generated chaotic output sounds similar to the original when played as audio.

Similar periodic chaos might be achievable using conventional supervised learning with RNNs.
However, it requires minimizing the prediction error while ensuring $\mathrm{MLE}>0$, leading to a significantly complex design and computation of the loss function \cite{mikhaeil2022difficulty}.

One application of the method is in the control of systems, including robotics. 
For example, chaos can be utilized in robot motor commands to avoid deadlock or to escape when a limb becomes trapped in a confined space \cite{steingrube2010self}. 
In such cases, the employed chaotic trajectories should be designed with consideration for the robot's body morphology and desired locomotion pattern.
Additionally, periodic chaos can aid in synchronizing nonlinear systems.
Carroll \textit{et al.} \cite{carroll1993using} discussed systems driven by periodic inputs that behave with $n$-multiple periods of the driving signal.
Their phase depends on initial conditions, which is problematic when driving multiple systems with the same input to achieve synchronized $n$-period behavior, such as the coordination of robot parts that employ nonlinear materials.
According to the authors, this issue can be resolved by using pseudo-periodic inputs with chaotic or stochastic variations instead of periodic inputs. Our method can help generate these pseudo-periodic inputs, which resemble the original input.

Finally, although this paper focuses on a method employing an LESN, future work could extend to physical reservoir computing \cite{nakajima2020physical}. 
This is particularly relevant for autonomous control systems that expect both appropriate behavior and emergent properties \cite{akashi2024embedding}. 
In such cases, a control framework leveraging the system's intrinsic dynamical properties can be crucial.

\vskip\baselineskip
\begin{acknowledgements}
    K. A. was supported by 
    Moonshot R\&D Grant Number JPMJMS2021,
    the institute of AI and Beyond of UTokyo,
    the International Research Center for Neurointelligence (WPI-IRCN) at The University of Tokyo Institutes for Advanced Study (UTIAS),
     JSPS KAKENHI Grant Number JP20H05921,
    Cross-ministerial Strategic Innovation Promotion Program (SIP), and the 3rd period of the SIP ``Smart energy management system" Grant Number JPJ012207
\end{acknowledgements}

\bibliography{liter}
\end{document}



\title{Supplementary Material for\\
    ``Designing Chaotic Attractors: A Semi-supervised Approach''}

\author{Tempei Kabayama}
\author{Yasuo Kuniyoshi}
\affiliation{%
 Graduate School of Information Science and Technology, University of Tokyo, Bunkyo-ku, Tokyo 113-0033, Japan
}%

\author{Kazuyuki Aihara}
\affiliation{
 International Research Center for Neurointelligence, University of Tokyo Institutes for Advanced Study, University of Tokyo, Bunkyo-ku, Tokyo 113-0033, Japan
}%

\author{Kohei Nakajima}
\affiliation{%
 Graduate School of Information Science and Technology, University of Tokyo, Bunkyo-ku, Tokyo 113-0033, Japan
}

\date{\today}

\maketitle

The supplemental material contains:
\begin{itemize}
   \item Information on the relationship between effective spectral radius \cite{jaeger2007optimization} and maximum Lyapunov exponent (MLE) of a leaky-integrator echo-state network (LESN)
   \item Additional figures on the properties of reservoir computing (RC) systems trained on periodic orbits
   \item Analysis of the effects of washout length near the edge of chaos
\end{itemize}

\section{Effective Spectral Radius and Dynamical Properties}
\begin{figure}[htbp]
    \begin{center}
     \includegraphics[scale=0.9]{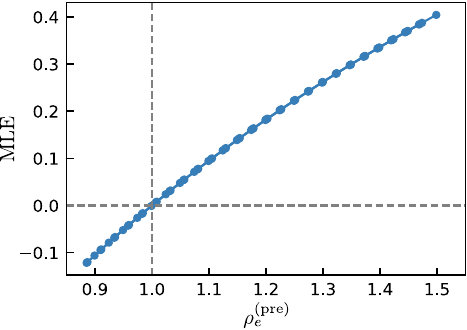}
     \caption{
        The effective spectral radius $\rho_e^{(\mathrm{pre})}$ and the MLE, 
        overlaying the plots for three different realizations of $\bm{W}$.
     }
     \label{fig:base}
    \end{center}
\end{figure}

An input-free LESN is a random recurrent neural network defined as follows:
\begin{equation}
    \bm{x}_{k+1} = (1-a)\bm{x}_{k} + a \tanh(\rho \bm{W} \bm{x}_{k}), \label{eq:lesn}
\end{equation}
where $\bm{x}$ is an $N$-dimensional state vector.
Its effective spectral radius is described by
\begin{equation}
    \rho_{e}^{(\mathrm{pre})} = |\lambda|_{\mathrm{max}}(a \rho \bm{W} + (1-a) \bm{I}).
\end{equation}
Here, $\rho_e^{(\mathrm{pre})}$ is the absolute value of the eigenvalue of origin, the equilibrium point. 
Therefore, the origin is stable when $\rho_e^{(\mathrm{pre})}$ is smaller than unity.
Figure~\ref{fig:base} shows the relationship between $\rho_e^{(\mathrm{pre})}$ and the MLE in the range $\rho\in[0.8, 2.0]$ with $(N, a) = (1000, 0.5)$, overlaying the plots for three realizations of $\bm{W}$.
Note that $\rho_{e}^{(\mathrm{pre})} \simeq a \rho + (1-a)$ in general.
After destabilization of the origin, the attractor evolves into chaos, and its MLE is determined by $\rho_e^{(\mathrm{pre})}$.

\begin{figure}[htbp]
    \begin{center}
     \includegraphics[scale=0.9]{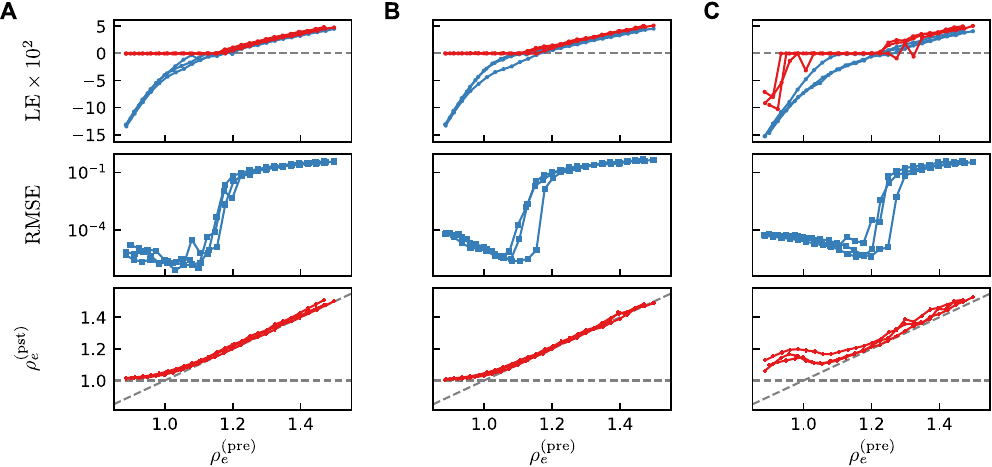}
     \caption{
     Properties of the driven and trained systems when using Lissajous curves (A), a unit circle (B), and Van der Pol oscillators (C) as the \textit{skeleton}. 
     Fixing $(N, a, \sigma) = (1000, 0.5, 0.2)$, the results for three different realizations of the random matrix pair $(\bm{W}, \bm{W}_{in})$ are overlaid.
     }
     \label{fig:prop}
    \end{center}
\end{figure}

Multiple quantities of driven LESNs and trained ones are shown in Fig.~\ref{fig:prop}, with respect to the \textit{skeleton} $\bm{u}_k$ of a Lissajous curve (Fig.~\ref{fig:prop}A), a unit circle (Fig.~\ref{fig:prop}B), and a Van-der-Pol oscillator (Fig.~\ref{fig:prop}C), overlaying the results for the three different realizations of  $(\bm{W}, \bm{W}_{in})$.
Here, the following properties are commonly observed even when the realizations of the random matrices or the \textit{skeleton} are altered:
\begin{itemize}
    \item The CLE is strongly determined by $\rho_e^{(\mathrm{pre})}$.
    \item Excessively slow convergence negatively impacts the prediction tasks.
    \item $\rho_e^{(\mathrm{pst})}$ adheres to $\rho_e^{(\mathrm{pre})}$ near the edge of chaos.
\end{itemize}
In the case of the Van der Pol oscillator, successful embedding is achieved only for relatively large values of $\rho_e^{(\mathrm{pre})}$. For $\rho_e^{(\mathrm{pre})}<1.0$, trajectories corresponding to the \textit{skeleton} become destabilized, and untrained attractors (such as fixed points, periodic orbits, and chaotic attractors) emerge, dominating the phase space. The instability of the MLE shown in the upper plot of Fig.~\ref{fig:prop}C stems from the presence of these untrained attractors.

\section{Bifurcation structures around semi-supervised points}
We overview the bifurcation structures around the semi-supervised points $\rho^{*}$.
In the main text, we plot the extrema of the node averages $\bar{x}^i$ to create bifurcation diagrams for an $N$-dimensional dynamical system that behaves like continuous time.
This method is simple and easy to define. 
However, it has drawbacks, such as the fact that the state variable $\bar{x}^i$ often takes a small values due to the cancellation of positive and negative values, potentially obscuring fine bifurcation structures. 
Therefore, focusing on a particular node $x^i$ can be effective. 
Additionally, instead of choosing extrema, one could define an $N-1$ dimensional hyperplane and observe the resulting Poincar\'{e} map. 
In RC systems, the output $\bm{z}_k$ can be utilized to select the Poincar\'{e} section. For example, the $x$-axis in the output space corresponds to the hyperplane $z^{(y)}=\bm{w}_y^\top \bm{x}=0$. 

\begin{figure}[htbp]
    \begin{center}
     \includegraphics[scale=0.9]{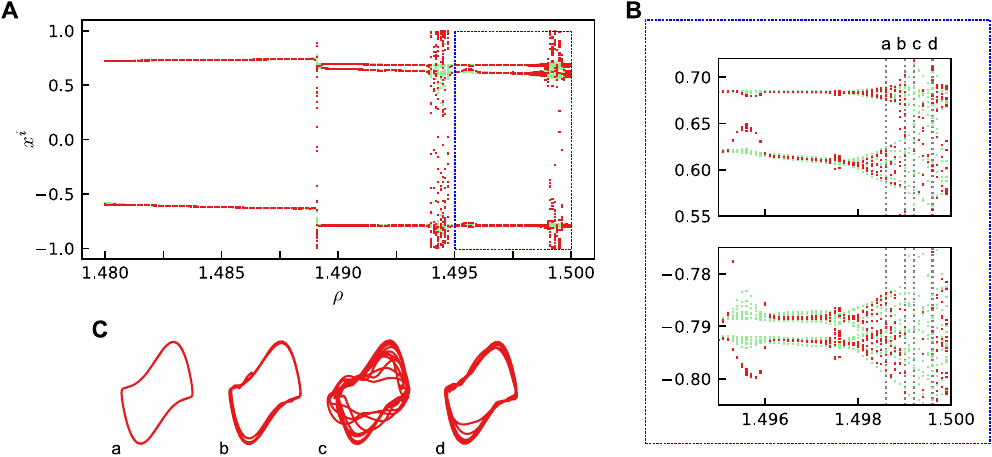} 
     \vspace*{0.25cm}
     \caption{
     (A) Bifurcation diagram of the Poincar\'{e} map at $z^{(y)}=0$ for a particular node $x^i$ in the region $\rho \in [1.48, 1.50]$ around the semi-supervised points when the Van der Pol oscillator is used as the \textit{skeleton}. Out of a total calculation for 10,000 steps, the first 2,000 steps are shown in light green, and the last 2,000 steps are shown in red.
     (B) Enlarged view of $\rho \in [1.48, 1.50]$. 
     (C) Output at $\rho \in \{1.4984, 1.4990, 1.4992, 1.4996\}$ indicated by the dotted lines a--d in B.
     }
     \label{fig:bif_vdp}
    \end{center}
\end{figure}

Figure~\ref{fig:bif_vdp} shows a bifurcation diagram with a Van der Pol oscillator as the \textit{skeleton} for the region around the semi-supervised points $\rho \in [1.48, 1.50]$, which is obtained by focusing on a particular node $x^i$ and plotting the Poincar\'{e} map based on the hyperplane $z^{(y)}=0$. 
In Fig.~\ref{fig:bif_vdp}A, changes in the stability of periodic orbits and multiple crises are observed. 
Furthermore, the enlarged view for the region $\rho \in [1.495, 1.500]$ in Fig.~\ref{fig:bif_vdp}B shows a cascade of period-doubling bifurcations leading to the semi-supervised points. 
Beyond the end of the cascade, the points maintaining periodic chaos coexist with those disrupted by crises. At $\rho=1.4984$, an output almost identical to the \textit{skeleton} is obtained (Fig.~\ref{fig:bif_vdp}Ca), while at $\rho=1.4990$ and $\rho=1.4996$, the output with some width of the trajectory is observed (Fig.~\ref{fig:bif_vdp}Cb and \ref{fig:bif_vdp}Cd). At $\rho=1.4992$, the periodic chaos is disrupted (Fig.~\ref{fig:bif_vdp}Cc).

\begin{figure}[htbp]
    \begin{center}
     \includegraphics[scale=0.9]{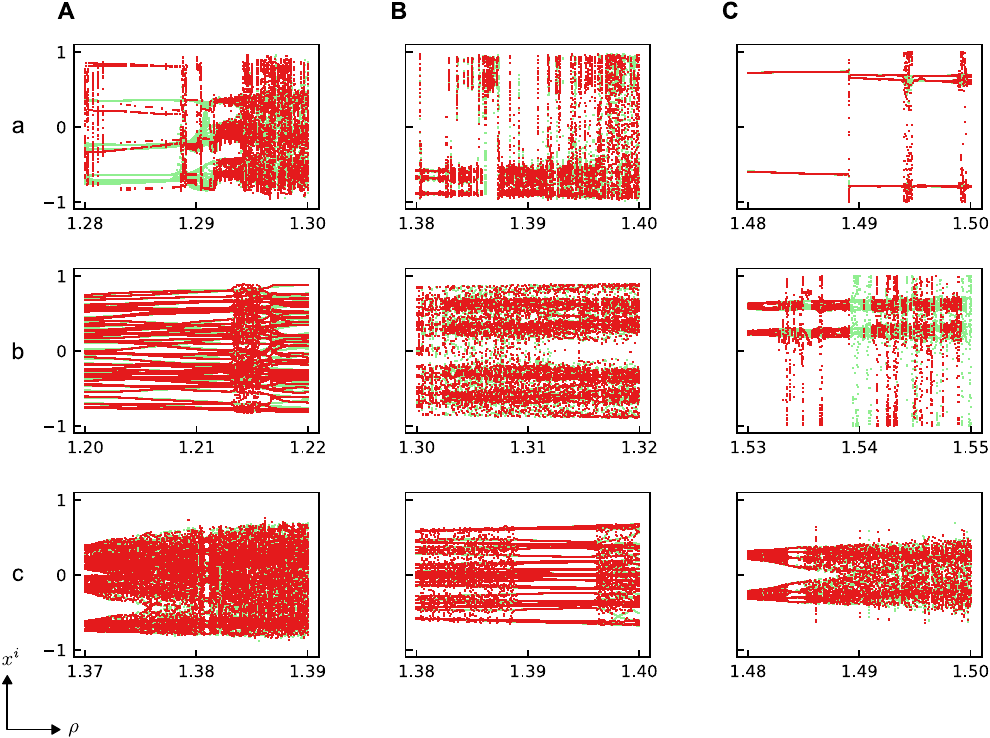}
     \caption{
     Bifurcation diagrams around the semi-supervised points for each setting with Lissajous curves (A), a unit circle (B), and Van der Pol oscillators (C), using three different realizations of the random matrix pair $(\bm{W}, \bm{W}_{in})$ (a--c).
     }
     \label{fig:bif_all}
    \end{center}
\end{figure}

Figure~\ref{fig:bif_all} shows
bifurcation diagrams around semi-supervised points, created using the same plotting method, for the nine settings consisting of three \textit{skeletons} (labeled A--C) and three realizations of matrices (labeled a--c) in Fig.~\ref{fig:prop}.
Although the appearance of the diagrams vary, the band regions consistently appear within some characteristic bifurcation structures like several periodic windows, period-doubling bifurcations, and crises.

\begin{figure}[htbp]
    \begin{center}
     \includegraphics[scale=0.9]{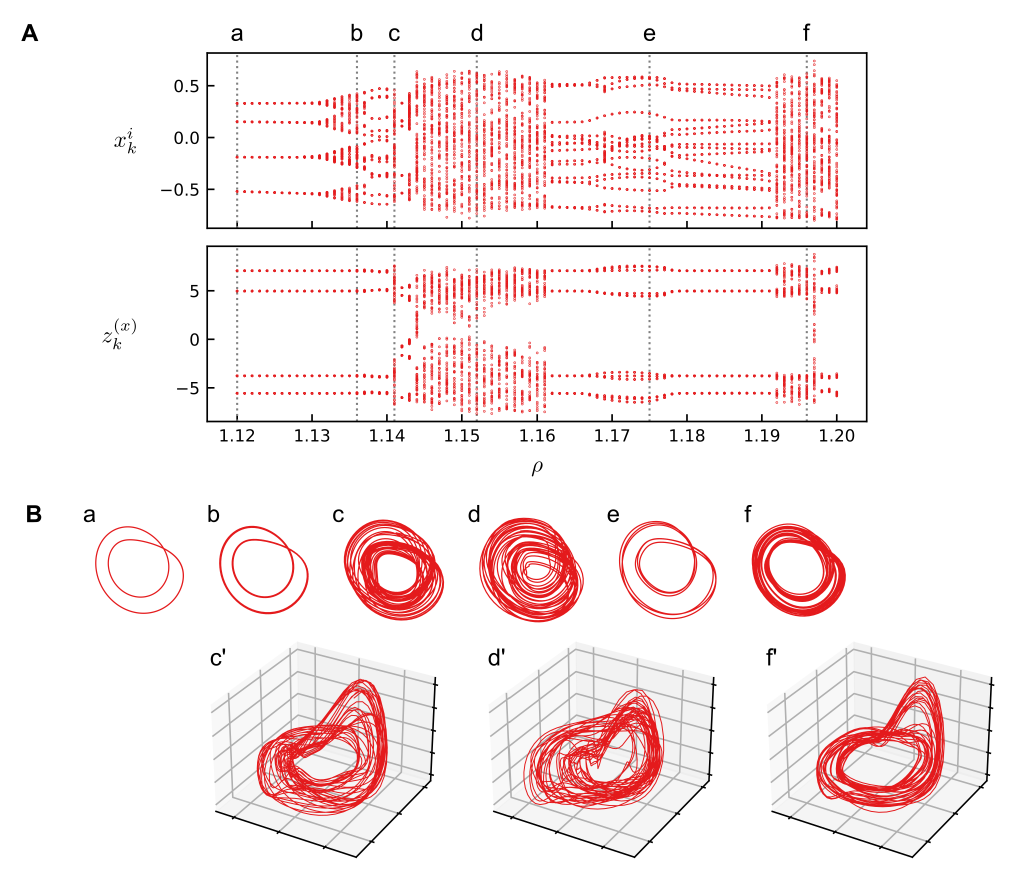}
     \caption{
     (A) Bifurcation diagram against $\rho$, created by plotting extrema. 
     The top shows a certain internal node, while the bottom shows the x-component of the output.
     (B) Plots in the $x-y$ plane of the output at each point $\rho\in\{1.120, 1.136, 1.141, 1.152, 1.175, 1.196\}$ indicated by the dotted lines a--f, and three-dimensional plots at points c, d, and f.
     }
     \label{fig:ros}
    \end{center}
\end{figure}

Interesting results can be obtained when using the periodic orbits of the R\"{o}ssler system as the \textit{skeleton}:
\begin{equation}\label{eq:ros}
    \begin{aligned}
    \dot{x} &= -y-z,\\
    \dot{y} &= x + 0.2y,\\
    \dot{z} &= 0.2 + xy - cz,
    \end{aligned}
\end{equation}
where $c=3.0$.
Figure~\ref{fig:ros}A shows the bifurcation structure from $\rho^{(\mathrm{P})}=1.12$ to $\rho^{(\mathrm{E})}=1.20$.
Here, in addition to the attractors with periodic chaos that closely fit the \textit{skeleton} (Fig.~\ref{fig:ros}Bb), we can also observe several chaotic attractors with expanded regions along the \textit{skeleton} (Figs.~\ref{fig:ros}Bc, \ref{fig:ros}Bd, and \ref{fig:ros}Bf).
These have the appearance of a M\"{o}bius strip, resembling the strange attractors of the original system \cite{rossler1976equation}, although they are not exactly the same.
The R\"{o}ssler system reaches chaos through a period-doubling route, for example, when increasing parameter $c$ in Eq. \eqref{eq:ros} \cite{stogatz1994nonlinear}, and it is speculated that similar period-doubling structures are involved in the emergence of periodic chaos in RC systems in general.
It is likely that the sharing of the route leading to chaos is the reason the chaos which was generated in the RC system has structures similar to those in the original system.

\section{Power spectrum analysis of the obtained time series}
\begin{figure}[htbp]
    \begin{center}
     \includegraphics[scale=0.9]{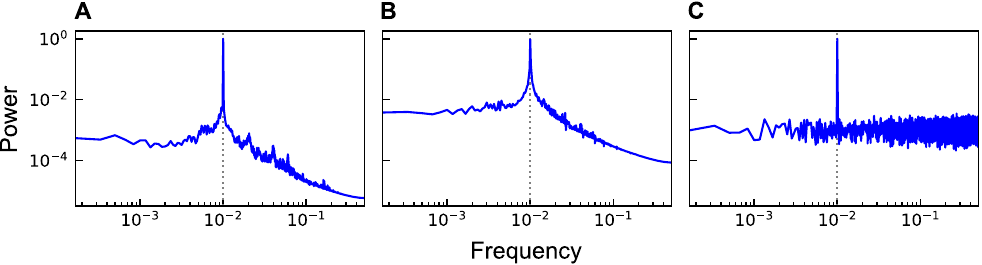}
     \caption{
     Power spectrum analysis results
     (A) for the chaotic orbits obtained at $\rho=1.2940$,
     (B) for the chaotic orbits obtained at $\rho=1.2995$,
     and (C) for the time series of the \textit{skeleton} with added Gaussian noise of variance 0.05.
     }
     \label{fig:psd}
    \end{center}
\end{figure}

To capture the characteristics of the chaotic trajectory obtained with the proposed method, power spectrum analysis was performed on the results with a Lissajous curve. 
Figures~\ref{fig:psd}A and \ref{fig:psd}B show the power spectra of the x-component of the orbits obtained at $\rho=1.2940 \leq \rho^{(\mathrm{E})}$ and $\rho=1.2995 > \rho^{(\mathrm{E})}$, respectively.
As expected, peaks can be observed corresponding to the 100-step period, but in addition to this, a flatter distribution is seen in the lower frequency region.
In comparison, Fig.~\ref{fig:psd}C shows the power spectrum when Gaussian noise with a variance of 0.05 is added to the \textit{skeleton}.
The power spectrum of the generated chaotic orbits exhibits a distribution like white noise in the low-frequency range, but qualitative differences can be observed in the high-frequency components. 
Additionally, it can be seen that the intensity of the low-frequency range is larger in an orbit with $\rho>\rho^{(\mathrm{E})}$ compared to an orbit with $\rho<\rho^{(\mathrm{E})}$.

\section{Influence of washout length at the edge of chaos}
At the edge of chaos, the driven system requires a significant number of steps to converge.
Therefore, within the fixed $T_{\mathrm{init}}$ steps, the transient may not fully vanish. Consequently, the training data $\bm{X}$ may contain information from the transient phase.
This information can affect the stability of the closed-loop system, both positively and negatively.
\begin{figure}[htbp]
    \begin{center}
     \includegraphics[scale=0.9]{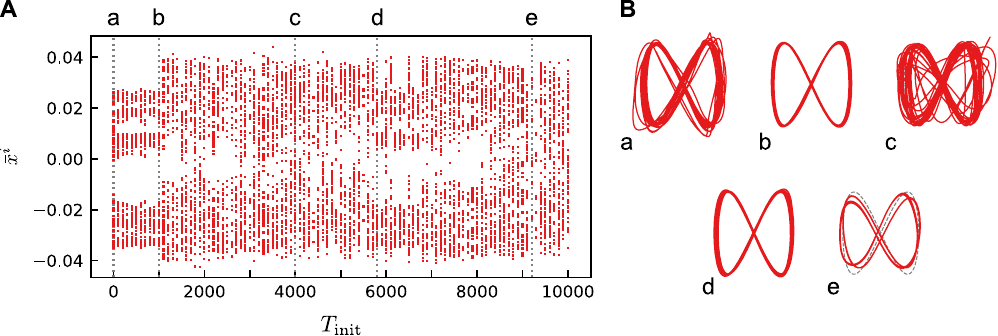}
     \caption{
     (A) Bifurcation diagram of the autonomous system trained by a Lissajous curves at $\rho=1.294$, varying with $T_{\mathrm{init}}$. (B) Output at $T_{\mathrm{init}}\in\{0,\, 1{,}000,\, 4{,}000,\, 5{,}800,\, 9{,}300\}$ indicated by the dotted lines a--f in A.
     }
     \label{fig:bif_wo}
    \end{center}
\end{figure}

When a Lissajous curve is used as the \textit{skeleton}, at $\rho=1.294$, the $\mathrm{CLE} \simeq 0$, and the convergence of the driven system is extremely slow.
At this point, by fixing $T_{\mathrm{train}}=2,000$ and varying $T_{\mathrm{init}}$ in increments of 100 steps, the bifurcation diagram in Fig.~\ref{fig:bif_wo}A is obtained. 
Here, regions with bands (Figs.~\ref{fig:bif_wo}Bb and \ref{fig:bif_wo}Bd) coexist with disrupted regions (Figs.~\ref{fig:bif_wo}Ba and \ref{fig:bif_wo}Bc), and short periodic windows can also be observed (Figs.~\ref{fig:bif_wo}Be).
In such regions where convergence is extremely slow, $T_{\mathrm{init}}$ acts as a bifurcation parameter, and the bifurcation structure is quite complex.
For practical purposes, it is advisable to fix $T_{\mathrm{init}}$ at a suitable value and focus on regulating $\rho$.

\bibliography{liter_supp}